# ROBOT HUMAN INTERFACE FOR HOUSEKEEPER ROBOT WITH WIRELESS CAPABILITIES

Suhad Faisal Behadili

Computer Science Department, Science Collage, University Of Baghdad, Iraq

## ABSTRACT

*This paper presents the design and implementation of a Human Interface for a housekeeper robot. It bases on the idea of making the robot understand the human needs without making the human go through the details of robots work, for example, the way that the robot implements the work or the method that the robot uses to plan the path in order to reach the work area. The interface commands based on idioms of the natural human language and designed in a manner that the user gives the robot several commands with their execution date/time. As a result, the robot has a list of tasks to be doneon certain dates/times. However, the robot performs the tasks assigned to it without any human intervention and then gives feedback to the human about each task progress in a dedicated list. As well as, the user decides to get the feedback either through the interface, through the wireless communication, or both of them. Hence, the user's presence not necessary during the robot tasks execution.*

## KEYWORDS

*Housekeeper, robot, wireless, human, mobile, tasks.*

## 1. INTRODUCTION

In the research domain, it is now trivial to witness humanoid robots strolling around with biped legs, and some of them are fit for utilizing instruments intended for humans such as knives and sweepers. In the commercial environments, robots are now not restricted to industrial environments, but they are getting into home environments such as vacuuming robots [1, 2].

It is fundamental for robots of the new generation to meets some benchmarks. To begin with, these robots must be animated, implying that they ought to respond to the states evolving in their surroundings [3]. This requires a nearby coupling about recognition and movement. Second, personal robots ought to be adaptable to various users, and diverse physical environments. This typically requires reasoning and learning capabilities. Finally, robots should be accessible, meaning that they are able to explain their beliefs, motivations, and intentions at the same time, they should be easy to command and instruct, which requires some kind of robot-human interaction [4, 2].

Many proposed robot-human interfaces suggest different types of interaction between the robot and the human. Some examples of them are a command language interface system to operate a robot where the user uses already existing programming languages to control the robot [5], a human interface system operates a robot by hand gestures where the arm direction controls the robot movement [6, 7]. Also, a graphical human interface system operates a robot, where the interface displays an image being captured by the robot camera [3, 5]. In addition, the human controls the robot based on the picture displayed and using the graphical commands that are exists in the interface [8, 9]. As well as a user interface based on a video game interface and offers a valid roadmap for the environment. The user controls the robot just as he is playing a game [10].

Whereas, the human controls the robot movement and/or tasks in a condition, where the robot and the human are present at the same time. Almost works of the robot-human interaction area assumed that the human should be present at the moment that the robot does its task or at least in the moment before the robot starts the task. That seems to conflate with the fact of freeing the human from doing the task via using the robot, and to have more time to do other things.

Besides, a vast majority of work in this domain incorporates convoluted graphical/command interfaces, wherein the user needs to learn not just the interface, but also the exhaustive robotic system hardware and software. For example, the human should specify the way of task implementation to the robot like which path the robot walks through to reach its goal, and which actions to be done in order to complete the specified task. This complicates human tasks instead of making human's life easier, and it requires a long time for learning how to make the robot do what it has to do.

Housekeeper robots have several features that are provided by the manufacture including: The user does not need to program the robot. This feature provided for home robots only to free the normal user from the complication of robot programming. Whereas, for industrial robots, the manufacturing companies allows engineers to program the robot [4]. The robot contains a database for the most common actions/responses to different situations in human's life, and it contains commands and algorithms that enable it to understand and respond to the user commands such as preparing a meal or washing dishes. However, the database helps the robot to perform its housekeeping responsibilities [5, 10]. The robot implicates path planning algorithms that are essential for the robot mobility within the surroundings [3], and the user does not have to be concerned in robot path planning [1]. Also, the robot has image processing capabilities to detect changes that happen to its surrounding, and to monitor events that occur in the house. This requires a camera existence, which is normally positioned in the robot head. The cameras also exist in other places like robot back, robot chest, room wall, etc. [8, 9].The robot contains artificial intelligence (AI) program that enables it to learn and adopt the new environments that are adapted on daily basis because the robot works in a very changed environment, which related to real human life. The AI algorithm updates the robot database, this producing more humanized robot behavior [4].

The above features are very important for any robot to be capable of working efficiently in a human's environment. In this work, the designed robot-human interface supports all the previous features plus the capability of monitoring and controlling the robot in order to learn and understand in easy manner. They will be discussed later in this paper, although they are not part of the designed interface.

In this investigation a human interface for a housekeeper robot based on idioms of the natural human language. The robot interface has several responsibilities inside the house including monitoring a small child, feeding him, playing with him, and preparing food/drinks like breakfast, launch, cafe, juice, etc. In addition, cleaning the house and washing the platters. The user does not have to be existent while the robot accomplishes the task. Additionally, it programs the robot to do different tasks in different times, and the robot does these tasks as scheduled. As well as, maintains the wireless communication between the robot and the user. The wireless communication becomes part of human life that enables human to communicate with others through mobile phones and/or the internet. Hence, the robot and the user could communicate through a mobile phone [3], while the user is outside the house. This provides the user with monitoring/controlling capabilities on the tasks that are going on in the house during his absence [2].

## 2. THE ROBOT HUMAN INTERFACES

As concerning any robot to have the place in a human environment, it has some sort of artificial intelligence program that enables the robot to better understanding the surroundings, and most important is to understand people that live in that surrounding. The AI itself is not part of our work, but we give the robot in our interface the capabilities of adding new tasks, new SMSs information, and/or new SMSs lists of todo actions.

The proposed robot-human interface consists of several pages allowing the user to define the robot tasks. Although, the interface based on graphical design, it uses natural human language to define the tasks. For example, if we want the robot to play with the baby, we select a task named "Play with the baby".

The interface designed based on the object-oriented philosophy where each page supports a certain object. The objects in this case, are the features that the housekeeper robot has. These features and the related interface pages discussed in the following sections.

In this work, the user starts adding tasks to the robot through its interfaces. The user selects a task from a task list which the robot able to execute. The user specifies the task date, time, and priority. Then adds the task to the temporary task list. Whenever the user finishes adding all tasks, so they are added to the "to do list", and the robot starts executing them. The robot implements a task until a higher priority task occurs. Hence, the robot postponed the current task and starts implementing the new one of higher priority.

Although the robot monitors the house all times, however, the user can add a task to the robot named monitor the house, so adding this task indicates that the user expects to go out at that time. Accordingly, the robot should be ready to start wireless communication with the user. The SMS sending procedure starts like a new task procedure for an event to occur in the house. Consequently, the robot checks if the event is emergency or not. So that, in emergency cases the robot follows the emergency predetermined procedure, then sends SMS informs the user. Otherwise, if the event is not anemergency, then the robot searches its database for SMS which the user prefers to receive from the interface, and if he chooses to receive the current event SMS, then the robot sends to him an SMS containing the occurred event with its available solutions. As well as, the robot waits 2 to 5 minutes to receive the user replay depending on the event itself. Thereafter, if the user for any reason (including mobile coverage problems, wireless communication limitations, not being fast enough to send a reply, or busy in doing another thing) doesn't reply to the SMS. Accordingly, the robot starts the default procedure for the selected event by the user through the interface.

Actually, the moving objects are monitored by the robot via several cameras. Hence, if the robot detects someone outdoors, then compares the face image of this person with the available images in the interface. Thereafter, if the robot identifies the person outside the door, then informs the user about this person. Otherwise, it informs the user about an unidentified person's existence outdoors. Finally, plans its path inside the house based on the house map using the famous Dijkstra algorithm.

## 3. TASK SCHEDULING

The robot performs several tasks in the house. The human himself assigns them; he specifies the needed tasks to be done along with the date and time in order to start its execution. The used approach constructs a list of "to do task", the robot have the scheduled tasks to be done. The user adds one or more tasks at one time, even if they occur at different times during the day, or on different dates.

However, some tasks take longer time than the user estimated, and then some urgent tasks could occur in the house with no previous planning, hence the user prioritizes the tasks for the robot. Then, the robot executes the scheduled task until its ending, a higher priority task emerges in the house, or it is already in the scheduled task list.

The house owner adds tasks for the robot using "New Tasks" page. Therefore, determines the date and time for the robot to do the task, and then determines the task type, and its priority. However, tasks with lower priorities would be delayed, whenever the tasks of higher priorities are in progress as shown in Fig (1). As well as, the user monitors the under execution task and the progress of the scheduled tasks (queued, in progress, done, and/or failed) through the "Current tasks" page. Therefore, the failed tasks have a reason of the failure in the "Progress" field of the "Current Tasks" page as shown in Fig (2).

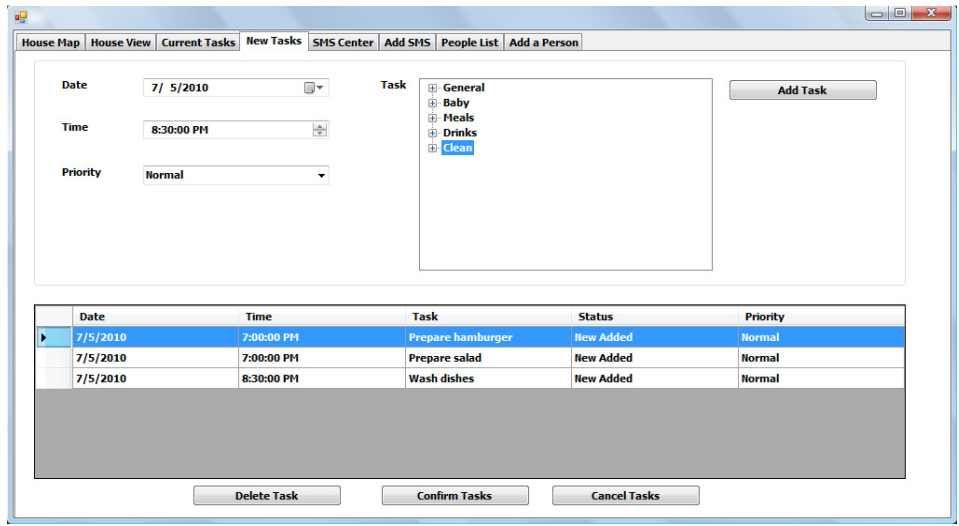

Figure 1. New Tasks Page

Figure 2. Current Tasks Page showing the statues of each task progress

## 4. PATH PLANNING

In pursuance of performing a robot task, it moves in around the house. Meanwhile, the robot should avoid collision with the surrounding environment. Hence, there are several methods for robot path planning. This field is still an important field for researchers. In this work, the robot uses a path planning method that complies with two facts. The first fact is the existence of the house map. Accordingly, the robot already knew the required paths in the house. The house map is loaded through our robot user interface, and it is stored to be used by the robot, and to be modified by the user when there is any necessity. The dimensions of the map and the positions of the house furniture that appeared on the map reflect the reality of the house as shown in Fig (3).

The second fact is the robot faces the moving obstacles during its moving. For example, a moving object/ human or repositioned furniture. Here, the interface uses a Grid-Based path planning methods named Dijkstra to divide the house map into small points called nodes. Consequently, each time the robot moves, then the method generates a list of nodes that the robot used to pass during its movement. Therefore, the house robots normally have a camera positioned in their head and/or sensors in different places of their bodies. Thus, they detect sudden appeared obstacles. Furthermore, while the data sent by the camera/sensors, then the path planning method (stored in the robot) regenerates a list of nodes, which the robot passes through to reach its target place. Commonly, the manufacturer provides the path plan method in real time to avoid obstacles. Meanwhile, in this work, the path plan method based on the house map is exclusive to this work.

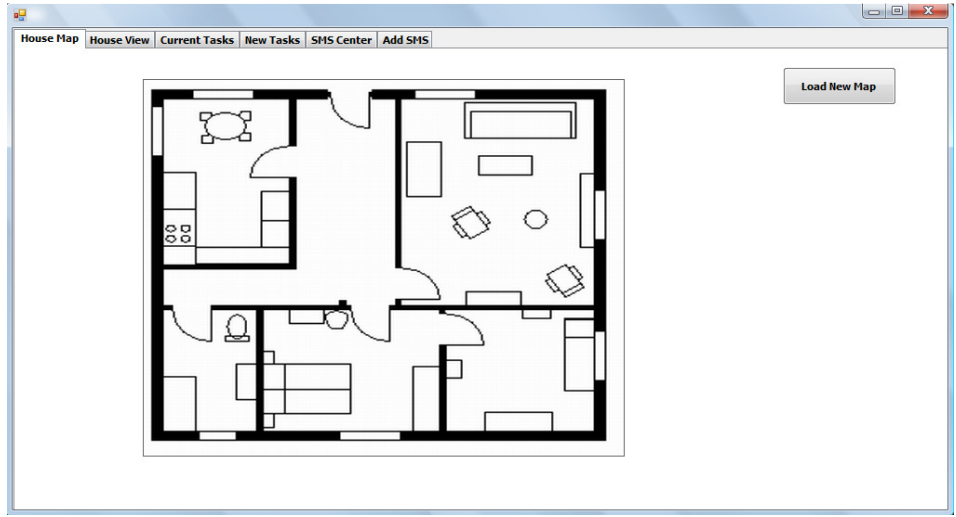

Figure 3. House Map Page

## 5. IMAGE POSSESSING

In this work, the robot has an image processor occupied internally in its system. As mentioned earlier. The robot recognizes moving objects that could become obstacles to the robot movement. Therefore, that means it must have an image detection processor. Moreover, these images obtained via a camera positioned in the robot head.

Besides the robot head camera, this interface supports the existence of several cameras positioned in all rooms, and in the outside door. Moreover, thanks to the existence of an image recognition processor, which supports the robot to determine moved furniture within the house immediately, and update the house map to reflect the new changes.

In addition, these cameras help the robot to monitor the house while nobody at home, so in case an emergency or unsuspected event happened, the robot takes action immediately. In addition, the user handles the images obtained from these cameras to monitor the house through the interface. In the house view page shown in Fig (4), the user sees four image shots for different rooms in the house. These images rotate every 30 second, so he sees a new room image each 30 second. In other words, he sees the living room, the kitchen, the baby room, and the main room in one 30 second. In another 30 second he sees the same first three rooms plus the outside door view, for example instead of the main room, and etc.

In this work, the robot has also a face detection capability, so it knows who is inside the house, and who is knocking the door. As in Fig (5), the user adds new people to the robot database through our robot human interface. He puts the person's name, image, telephone number, and mobile phone number. Whenever any person existed in the people list knocks the door or rings the phone, then the robot knows visiting/calling person, and tells the user these information. Thus, the list of confirmed people to the robot displayed through our interface of the people list page as shown in Fig (6).

International Journal of Computer Networks & Communications (IJCNC) Vol.10, No.3, May 2018

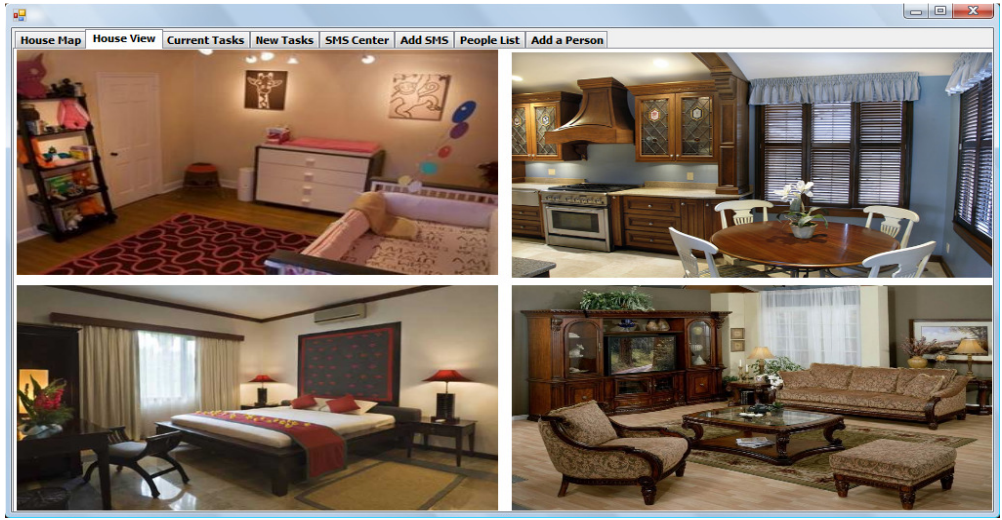

Figure 4. House View Page

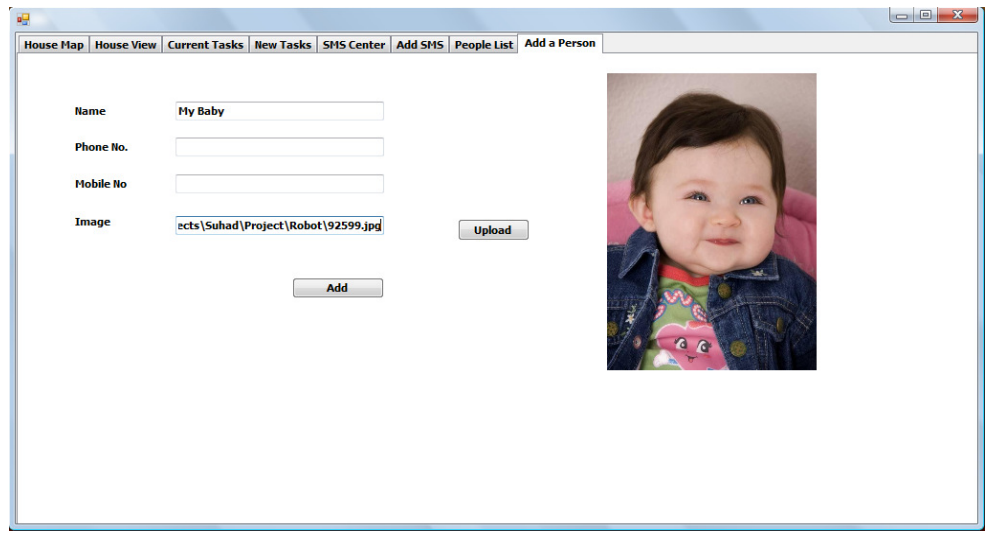

Figure 5. Add a New Person Page

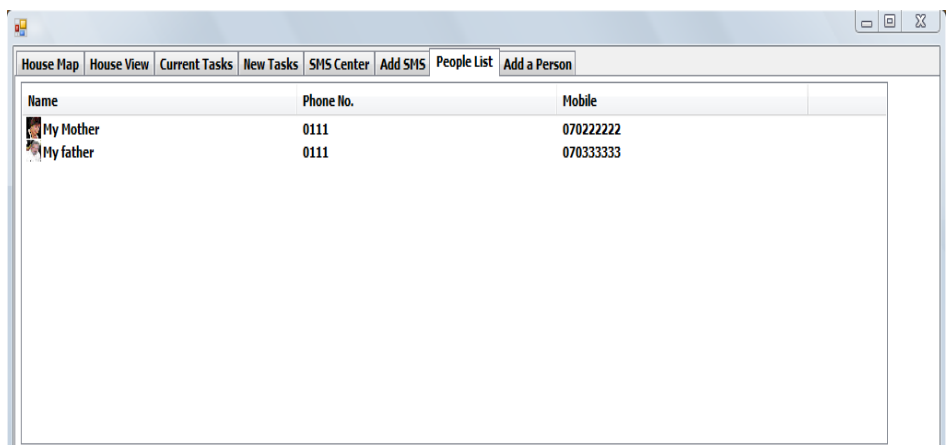

Figure 6. People List Page

## 6. WIRELESS COMMUNICATION

As mentioned earlier, the robot monitors the house while the user is outdoors. The robot also has a built-in mobile, which employs the wireless communication to make the user conscious of what is going on in the house during his absence [2]. The robot uses SMS messages to send this information to the user. There are two types of SMS messages, which our robot sends: The first type is an Emergency-SMS, the robot sends them once an emergency occurs in the house like a fire. The robot has a built-in list for these kinds of emergencies, so it knows exactly what to do if one of them happened. For example, if a fire breaks out, firstly it takes the baby outside the house, then calls the emergency number, and gives them the house address, the name of the owner, and the type of the occurred emergency, which is in our case a fire.

The second type of SMS messages, which the robot sends to the user, is the Reaction-SMS, which are the messages that need user reaction. For example, if your mother visits you while you are outside, the robot sends you an SMS to ask you if the robot lets her in, or if it simply takes a message from her. The SMS body contains two parts. The first part is the information it delivers to you. As well as, the second part is a numbered reactions list, which it considers, but it needs user interaction to choose one of them. He has to replay with the reaction number to the robot, where as he does not need to write the preferred commands completely. Back to the previous example, if letting the mother get in, so the reaction has a number equal to one, then the user sends just number one to the robot, so the robot opens the door, welcome the user's mother, and additionally serves her a juice.

The robot human interface enables the user to select the Reaction-SMS he wants. Note that Emergency-SMS delivered to the user always, and the user cannot choose not to receive them, the SMS page as in Fig (7)

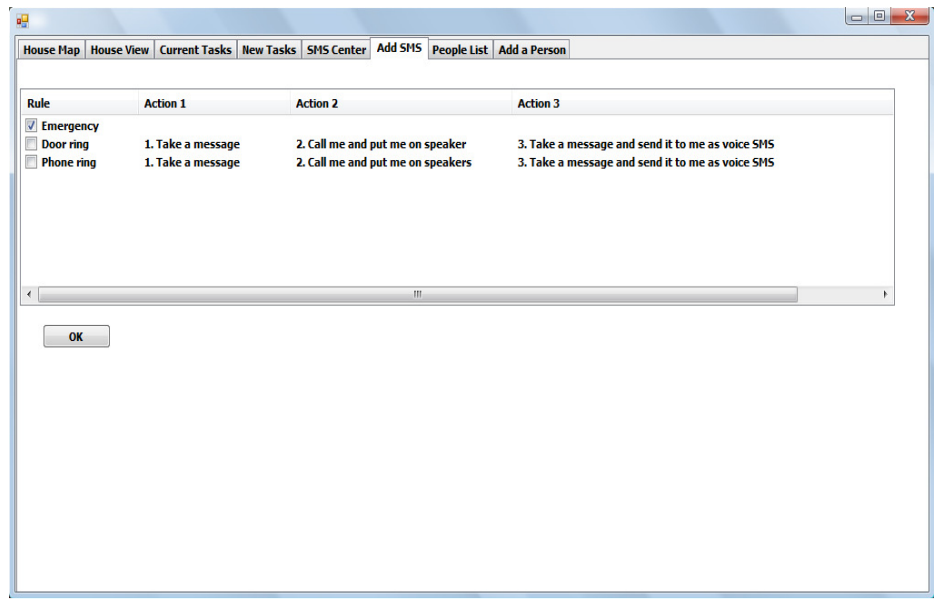

Figure 7.Add SMS Page

International Journal of Computer Networks & Communications (IJCNC) Vol.10, No.3, May 2018

The robot has also a default reaction, which the user determines for it in case that the user did not answer the robot SMS. The user interface allows the user to see all SMSs being sent by the robot, and the robot reaction either they based on a user replay or on a default reaction as shown in Fig (8).

Figure 8. SMS Center page

## 7. CONCLUSIONS

The discipline of robot-human interface is still needs many researchers' efforts, in order to implement easy interfaces to use by the human, and do not imply human teaching the robotics system details. This paper presents the robot- human interface for housekeeper robot. It assumes that the robot has many capabilities, which are supposed to be important for any robot to work in a human environment like a house. This interface designed in a way that enables the human to give scheduled tasks to the robot, and it implementing these tasks according to their schedule. The robot provides the resulted task implementation to the user via the interface and SMS messages. The SMS messages used in order to keep the user aware of tasks implementation in the house while the human is outdoors. The interface takes into consideration passing the feedbacks to the user, especially the interesting events for him, and he has to get their feedback. The interface design is simple and can used by any user. Also, covers almost information the robot may need from the human, like passing out the house map to it. Finally, the robot learning capabilities adapts the interface.

## REFERENCES

[1] Daisuke Sakamoto, Koichiro Honda, Masahiko Inami, and Takeo Igarashi, "Sketch and Run: A Stroke-based Interface for Home Robots", ACM, 2009.

[2] Shiu Kumar, "Ubiquitous Smart Home System Using Android Application", International Journal of Computer Networks & Communications (IJCNC), Vol.6, No.1, January 2014.

[3] Emi Mathews, and Ciby Mathew, "Deployment Of Mobile Routers Ensuring Coverage and Connectivity", International Journal of Computer Networks & Communications (IJCNC), Vol.4, No.1, January 2012.

[4] L. Seabra Lopes, A. Teixeira, D. Gomes, C. Teixeira, J. Girão, and N. Sénica, "A Friendly and Flexible Human-Robot Interface for CARL", Robotica 2003, Centro Cultural de Belém, Lisboa, 2003.

[5] JuliaBerg, and GuntherReinhart "An Integrated Planning and Programming System for Human-Robot-Cooperation", 50th CIRP Conference on Manufacturing Systems, Volume 63, 2017.

[6] MuneebImtiaz Ahmad, Omar Mubin, Joanne Orlando, "A Systematic Review of AdaptivityinHuman-Robot Interaction, Multimodal Technologies and Interact", 20 July 2017.

[7] Michael T. Rosenstein, Andrew H.Fagg, ShichaoOu, and Roderic A. Grupen, "User Intentions Funneled Through a Human Robot Interface", Proceedings of the 10th international conference on Intelligent user interfaces, 2005.

[8] Kazuhiko Kawamura, PhongchaiNilas, Kazuhiko Muguruma,Julie A. Adams, and Chen Zhou, "An Agent-Based Architecture for an Adaptive Human-Robot Interface", Proceedings of the 36th Hawaii International Conference on System Sciences (HICSS'03), IEEE, 2002.

[9] Carlotta A. Johnson, Julie A. Adams, and Kazuhiko Kawamura, "Evaluation of an Enhanced Human-Robot Interface", Proceedings of the IEEE International Conference on Systems Man and Cybernetics SMC IEEE, Oct. 2003.

[10] B. Maxwell, N. Ward, and F. Heckel, "Game-Based Design of Human-Robot Interfaces for Urban Search and Rescue", NSF IIS-0308186, and the American Association for Artificial Intelligence, In Proceedings of the CHI 2004 Conference on Human Factors in Computing Systems. ACM: The Hague, April 2004.

## AUTHOR

**Suhad Faisal Behadili** holds a PhD. of Computer Science/ networks and communicationsfrom LITIS, Le Havre, Normandy University, France. She is an academic member in computer science department, College of Science, Baghdad University. Her main interests are modeling and simulation.

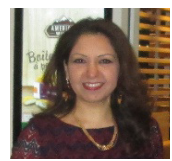